# A Cutting Mechanics-based Machine Learning Modeling Method to Discover Governing Equations of Machining Dynamics

Alisa Ren[a], Mason Ma[a,b], Jiajie Wu[a,b], Jaydeep Karandikar[c], Chris Tyler[c], Tony Shi[a,b], Tony Schmitz[b,c]

[a]*Manufacturing Intelligence Dynamics Laboratory, University of Tennessee Knoxville, Knoxville, TN 37996, USA*
[b]*Maching Tool Research Center, University of Tennessee Knoxville, Knoxville, TN 37996, USA*
[c]*Manufacturing Science Division, Oak Ridge National Laboratory, Oak Ridge, TN 37830, USA*

**Abstract**

This paper proposes a cutting mechanics-based machine learning (CMML) modeling method to discover governing equations of machining dynamics. The main idea of CMML design is to integrate existing physics in cutting mechanics and unknown physics in data to achieve automated model discovery, with the potential to advance machining modeling. Based on existing physics in cutting mechanics, CMML first establishes a general modeling structure governing machining dynamics, that is represented by a set of unknown differential algebraic equations. CMML can therefore achieve data-driven discovery of these unknown equations through effective cutting mechanics-based nonlinear learning function space design and discrete optimization-based learning algorithm. Experimentally verified time domain simulation of milling is used to validate the proposed modeling method. Numerical results show CMML can discover the exact milling dynamics models with process damping and edge force from noisy data. This indicates that CMML has the potential to be used for advancing machining modeling in practice with the development of effective metrology systems.

*Keywords*: machining dynamics; modeling; delayed differential equations; cutting mechanics; machine learning

## 1. Introduction

Machining dynamics studies the physical process of metal cutting due to the interaction of computer numerical control (CNC) machine, tool, and workpiece. In the machining process, dynamic tool-workpiece engagement induces dynamic cutting forces, that can further excite the structural dynamics of machine tool to vibrate at the tool tip point, and eventually form the material surface of workpiece [1]. As such, modeling machining dynamics generally needs to integrate two parts: the structural dynamics of machine tool and cutting forces of dynamic cutting process.

Significant modeling efforts on machining dynamics have been made in the past decades. A consensus in the research community is that machining dynamics can be mathematically described by second order, coupled, time delayed differential equations (DDEs) of motion. The first model of machining dynamics was developed in [2] as a single degree of freedom DDE for turning dynamics. It described that chatter, or self-excited vibration, is primarily caused by regenerative chip thickness due to time delayed effect. The pioneer research findings of stability laws for turning then were concurrently and independently presented in [3, 4]. This single degree of freedom DDE for orthogonal cutting, however, cannot describe multiple degrees of freedom cutting like milling. For milling, the first model was then developed in [5] where time periodicity and time delay were formulated to characterize the milling dynamics. In addition to the time delayed effect, it was observed that the stability increases at low spindle speeds. This is attributed to the process damping effect caused by the penetration of cutting edge into wavy cut surface [6]. The process damping for turning and milling was studied in [7-9] by adding a process damping force, that is dependent on the surface normal velocity, chip width, cutting speed, and empirical coefficients. Other force models like edge force and runout were also studied based on different cutting mechanics [10]. What's more, the receptance coupling substructure analysis approach, analytically joining substructure receptance of tool, holder, and spindle-machine, was proposed in [11, 12] to advance machining modeling for better structural dynamics predication. These models are experimentally verified and widely used in academia and industry.

However, due to the complexity of stability solutions of DDEs, most of existing physical models are developed in linear form. Nonlinear models are seldom explored. One representative nonlinear method is time domain simulation that can well capture the nonlinearities caused by tool jumping out of the cut, runout, and varying tool geometry [13-15]. A few other studies of nonlinearities on structural dynamics and cutting forces are summarized in [16]. On the other hand, for data-driven methods, machine learning (ML) models (generally nonlinear) have been applied to machining dynamics, including Bayesian learning [17-19] and neural networks [20]. However, those ML models are mainly designed to generate more accurate stability solutions based on existing DDEs and experimental data, rather than discovering governing equations of machining dynamics. These limitations of current physical and ML models motivate this study to design an integrated physics-based and data-driven modeling method, that can leverage existing physics in cutting mechanics to explore unknown physics in data, and therefore discover nonlinear governing equations of machining dynamics.

This paper presents the first attempt to develop a cutting mechanics-based machine learning (CMML) modeling method to discover governing equations for machining dynamics. Based on existing physics in cutting mechanics, CMML first establishes a general modeling structure through representing the traditional DDEs of machining dynamics by a set of unknown differential algebraic equations. Then, to discover these unknown differential algebraic equations, effective nonlinear learning function space design and discrete optimization-based learning algorithm are developed as the ML component of CMML. Experimentally verified time domain simulations of milling with considerations of nonlinear process damping and edge force are used to validate the proposed CMML modeling method. Numerical results demonstrate CMML can successfully discover the exact milling dynamics models from noisy data. This indicates the proposed CMML modeling method has the potential for advancing machining modeling in practice with the development of effective metrology systems.

This paper is organized as follows. Section 2 presents the proposed CMML modeling method. The simulation experiments, numerical results and analyses are given in Section 3. Section 4 concludes this paper.

## 2. Proposed Cutting Mechanics-based Machine Learning Modeling Method

The proposed CMML modeling method is technically presented based on two sequential components: Establish a general cutting mechanics-based modeling structure by first principles (Section 2.1); and discover governing equations of machining dynamics from data by ML algorithm design (Section 2.2).

### 2.1 Establish Cutting Mechanics-based Modeling Structure by First Principles

#### 2.1.1 Time delayed Effect and Dynamic Cutting Force

Metal cutting consists of various processes, including the most common turning, milling, and drilling, and followed by special operations such as boring, broaching, and shaping. Although the tool geometry and tool kinematics of these cutting operations may differ from each other, they share the same principles of cutting mechanics [21]. Two fundamental principles of cutting mechanics are introduced below: 1) time-delayed effect, and 2) dynamic cutting force generation and projection. They lay a solid foundation to establish a general modeling structure for the proposed CMML modeling method.

Without loss of generality, a simple case of two-dimensional orthogonal cutting is used to explain these two fundamental principles, as shown in Figure 1. The tool edge is orthogonal to the cutting speed $V$. A metal chip with an instantaneous chip thickness $h(t)$ is sheared away from the workpiece. Let the chip thickness direction be normal or radial direction, and the direction of cutting speed $V$ be tangential direction.

Time delayed effect determines the instantaneous chip thickness $h(t)$, as shown in Eq. (1).

$$h(t) = h + n(t - \tau) - n(t). \tag{1}$$

$h$ is the static chip thickness attributed to rigid body motion of the cutting tool, and $n(t - \tau) - n(t)$ is the dynamic chip thickness caused by time-delayed effect. $n(t - \tau)$ and $n(t)$ are the vibrations of tool along chip thickness direction at previous and current cuts. The time delay $\tau$ is dependent on the spindle speed and can be different for different cutting processes. For example, $\tau = 1/\Omega$ for turning where $\Omega$ (rps) is the spindle speed. The time delay can be explained as follows. As soon as the cutting tool encounters the workpiece, it starts to vibrate because of the mass and stiffness. This leaves a wavy surface on the workpiece, as shown in Figure 1. The vibrations contribute to generating variable cutting force, and the cutting force will again generate vibrations of the cutting tool. Due to periodic movement of the spindle, when the next vibrating cutting edge encounters the wavy surface left by the previous cut, the dynamic chip thickness presents.

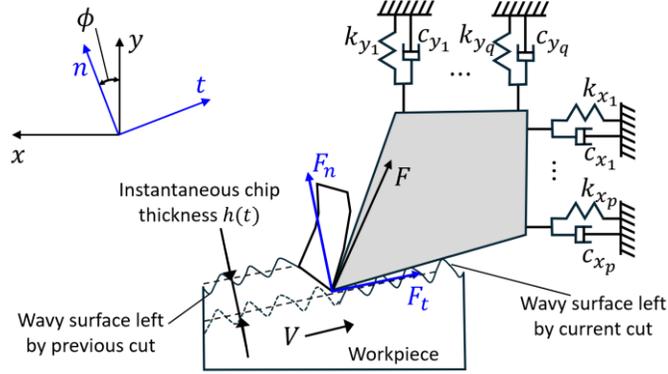

Figure 1: Schematic illustration of two fundamental principles of cutting mechanics: 1) time delayed effect, and 2) dynamic cutting force generation and projection. The resultant dynamic cutting forces excite structural dynamics to vibrate on machine tool's coordinate.

The resultant dynamic cutting forces are exerted on the tangential and normal directions, denoted by $F_t$ and $F_n$. In general, the cutting forces are functions that can be described as in Eq. (2):

$$F_t(t) = f_t(n(t - \tau) - n(t), \dot{n}(t), \boldsymbol{u}), \tag{2a}$$
$$F_n(t) = f_n(n(t - \tau) - n(t), \dot{n}(t), \boldsymbol{u}). \tag{2b}$$

$\dot{n}$ denotes the velocity of tool vibration on the normal direction, and vector $\boldsymbol{u} \in \mathbb{R}^{1 \times r}$ consists of $r$ cutting process parameters (e.g., spindle speed $\Omega$, cutting depth $b$, and feed rate $f_t$). The cutting forces $F_n$ and $F_t$ on the cutter's coordinate are projected onto the machine tool's coordinate as $F_x$ and $F_y$ to excite structural dynamics to vibrate on both directions. The projection relationship in Figure 1 can be resolved as

$$F_x = -F_t \cos(\phi) + F_n \sin(\phi), \tag{3a}$$
$$F_y = F_t \sin(\phi) + F_n \cos(\phi). \tag{3b}$$

$\phi$ is the angle between the chip thickness and machine tool's $y$ (normal) directions. In general, the projection angle $\phi$ can be a constant or time dependent variable. For example, in turning, this angle is a constant, while in milling this angle is a time varying rotational angle of cutting edge. The projection relationship in Eq. (3) is critical for studying structural dynamics as it shifts the dynamic cutting forces on the varying coordinate of tool to the fixed Cartesian coordinate of machine.

*2.1.2   Cutting Mechanics-based Modeling Structure Design*

The dynamic cutting forces on $x$ and $y$ directions of machine tool excite the structural dynamics on both directions to vibrate at the tooltip during machining processes. Thus, to model machining dynamics is to understand the sources of force generation and the resulting machine tool's structural dynamics excited by these forces. In the following, the structural dynamics of machine tool is assumed to have $p$ and $q$ modes on $x$ and $y$ directions, as shown in Figure 1.

In prior studies, the governing equations of motion for machining dynamics are commonly modeled by second order, coupled DDEs as shown in Eq. (4):

$$\boldsymbol{m}_x \ddot{\boldsymbol{x}}^T + \boldsymbol{c}_x \dot{\boldsymbol{x}}^T + \boldsymbol{k}_x \boldsymbol{x}^T = F_x(t, \boldsymbol{x}(t), \boldsymbol{x}(t-\tau), \dot{\boldsymbol{x}}), \quad (4a)$$
$$\boldsymbol{m}_y \ddot{\boldsymbol{y}}^T + \boldsymbol{c}_y \dot{\boldsymbol{y}}^T + \boldsymbol{k}_y \boldsymbol{y}^T = F_y(t, \boldsymbol{y}(t), \boldsymbol{y}(t-\tau), \dot{\boldsymbol{y}}), \quad (4b)$$

where $\boldsymbol{x}, \dot{\boldsymbol{x}}, \ddot{\boldsymbol{x}}, \boldsymbol{m}_x, \boldsymbol{c}_x$, and $\boldsymbol{k}_x$ are the displacement, velocity, acceleration, modal mass, viscous damping, and stiffness on $x$ direction and all vectors in $\mathbb{R}^{1 \times p}$; and $\boldsymbol{y}, \dot{\boldsymbol{y}}, \ddot{\boldsymbol{y}}, \boldsymbol{m}_y, \boldsymbol{c}_y$, and $\boldsymbol{k}_y$ are the displacement, velocity, acceleration, modal mass, viscous damping, and stiffness on $y$ direction and all vectors are in $\mathbb{R}^{1 \times q}$. The left-hand-side of Eq. (4) represents the structural dynamics of machine tool with respect to displacement, velocity, and acceleration. The right-hand-side of Eq. (4) is the cutting force model considering the time delayed term and projection. Common assumptions are imposed like orthogonality of $x$ and $y$ directions, linear model structure, and viscous damping [1]. These models are well studied and experimentally verified through decades of research efforts in the machining community.

Based on the time delayed effect and dynamic cutting force, a general cutting mechanics-based modeling structure is proposed in state-space form through representing the governing DDEs in Eq. (4) by a set of differential algebraic equations (DAEs) in Eq. (5) as follows.

First, the governing differential equations for displacement and velocity that describe the structural dynamics of machine tool can be formulated as the unknown equations in Eqs. (5a) – (5d):

$$\dot{x} = f_1(s(t), u, F(t)), \tag{5a}$$
$$\dot{v}_x = f_2(s(t), u, F(t)), \tag{5b}$$
$$\dot{y} = f_3(s(t), u, F(t)), \tag{5c}$$
$$\dot{v}_y = f_4(s(t), u, F(t)). \tag{5d}$$

Notations $v_x$ and $v_y$ are introduced by letting $v_x = \dot{x}$ and $v_y = \dot{y}$. Note that $s(t) \coloneqq [x(t)\ v_x(t)\ y(t)\ v_y(t)]$ in $\mathbb{R}^{1\times(2p+2q)}$ consists of all the state variables and $F(t) \coloneqq [F_x(t)\ F_y(t)]$ in $\mathbb{R}^{1\times 2}$ is the vector of cutting forces.

Then, by integrating the projection relationship from $F_n$ and $F_t$ to $F_x$ and $F_y$ in Eq. (3), the governing equations for cutting forces can be modeled as the following algebraic equations in Eqs. (5e) - (5h):

$$F_n = g_1(x(t), x(t-\tau), y(t), y(t-\tau), u), \tag{5e}$$
$$F_t = g_2(x(t), x(t-\tau), y(t), y(t-\tau), u), \tag{5f}$$
$$F_x = -F_t(t)\cos(\phi) + F_n(t)\sin(\phi), \tag{5g}$$
$$F_y = F_t(t)\sin(\phi) + F_n(t)\cos(\phi). \tag{5h}$$

It should be noted that, Eqs. (5e) and (5f) are unknown and once they are determined, the analytical forms of $F_x$ and $F_y$ in Eqs. (5g) and (5h) are determined. The general coupled dynamics on $x$ and $y$ directions are considered for both differential and algebraic equations. The coupled dynamics can be decoupled by assuming that zero cross talk is between the two orthogonal directions of machine tools.

Therefore, a general cutting mechanics-based modeling structure is obtained through representing the traditional governing equations of DDEs by a set of unknown DAEs, i.e., Eq. (5) including Eqs. (5a) – (5h). This general modeling structure for machining dynamics enables the possibility that utilizing existing physics in cutting mechanics to explore the unknown physics in data by developing effective machine learning algorithms to discover those unknown DAEs.

## 2.2 Discover DAEs from Data by Machine Learning

The discrete-optimization based machine learning method for discovering differential equations of nonlinear dynamics in [22] is extended as the ML component of the proposed CMML modeling method, to discover all the unknown DAEs.

*2.2.1 Cutting Mechanics-based Nonlinear Learning Function Space Design*

Throughout this paper, let $[P] = \{1, 2, \dots, P\}$ for any given positive integer $P$. In addition, for given process parameters $u$, the state variables $s(t)$, time derivatives $\dot{s}(t)$, and cutting forces $F(t)$ are assumed to be measurable in time series form. The resulting data matrices are denoted by $\mathbf{S} \in \mathbb{R}^{N\times(2p+2q)}$, $\dot{\mathbf{S}} \in$ $\mathbb{R}^{N\times(2p+2q)}$, and $\mathbf{F} \in \mathbb{R}^{N\times 2}$, respectively. This assumption is reasonable. For example, the dynamometers, capacitance probes, laser vibrometers, and laser accelerometers were used to concurrently measure cutting force, displacement, velocity and acceleration of cutting tools [23]. For ease of presentation, the process parameter vector $\boldsymbol{u}$ is rewritten in a matrix form $\mathbf{U} = \mathbf{I}_{N\times r} diag\{\boldsymbol{u}\} \in \mathbb{R}^{N\times r}$ with the same row number as $\mathbf{S}$. Herein, $\mathbf{I}_{N\times r} \in \mathbb{R}^{N\times r}$ is a matrix with all entries set to 1, and $diag\{\boldsymbol{u}\} \in \mathbb{R}^{r\times r}$ is a diagonal matrix with diagonal entries set to $\boldsymbol{u}$. In this way, $\boldsymbol{u}$ is constructed as paired data of $\boldsymbol{s}(t)$, $\dot{\boldsymbol{s}}(t)$, and $\boldsymbol{F}(t)$. The data in $\mathbf{S}$, $\dot{\mathbf{S}}$, and $\mathbf{F}$ can be injected with noises.

Note that the developed machine learning algorithm can be applied to all unknown DAEs in Eq. (5). Therefore, without loss of generality, the discovery of governing equation of displacement $x_j$ of the $j$-th mode is used throughout, to introduce the developed machine learning algorithm. The learning function space design for unknown DAEs is to create an augmented function space by expanding the function space that current DDEs-based governing equations live. Existing physical knowledge in cutting mechanics can be incorporated for the design of this expanded learning function space design.

The governing equation $\dot{x}_j = f_j(\boldsymbol{s}, \boldsymbol{u}, \boldsymbol{F})$ is to be discovered using data matrices $\mathbf{S}$, $\mathbf{U}$, $\mathbf{F}$, the process parameters $\boldsymbol{u}$, and the $j$-th column $\dot{x}_j$ of matrix $\dot{\mathbf{S}}$ for method illustration. The subscript $j$ is further omitted for notational simplicity. First, a cutting mechanics-based candidate set $\boldsymbol{\theta}(\boldsymbol{s}, \boldsymbol{u}, \boldsymbol{F})$ consisting of a total of $P$ candidate physical terms for $f(\boldsymbol{s}, \boldsymbol{u}, \boldsymbol{F})$ should be constructed as follows:

$$\boldsymbol{\theta}(\boldsymbol{s}, \boldsymbol{u}, \boldsymbol{F}) \coloneqq [\theta_1(\boldsymbol{s}, \boldsymbol{u}, \boldsymbol{F})\ \theta_2(\boldsymbol{s}, \boldsymbol{u}, \boldsymbol{F}) \cdots \theta_P(\boldsymbol{s}, \boldsymbol{u}, \boldsymbol{F})], \tag{6}$$

where $\theta_1(\boldsymbol{s}, \boldsymbol{u}, \boldsymbol{F}), \cdots, \theta_P(\boldsymbol{s}, \boldsymbol{u}, \boldsymbol{F})$ represent the candidate physical terms that can be contained in the unknown function $f(\boldsymbol{s}, \boldsymbol{u}, \boldsymbol{F})$. Existing physics in cutting mechanics needs to be incorporated into the construction of candidate set design. For example, in milling dynamics, besides the time delayed term and cutting force projection, the periodic immersion angle $\phi$ caused by the rotary milling cutter should also be taken into consideration. The corresponding cutting mechanics-based candidate set can be constructed as

$$\boldsymbol{\theta}(\boldsymbol{s}, \boldsymbol{u}, \boldsymbol{F}) = [1\ \boldsymbol{s}\ \boldsymbol{u}\ \boldsymbol{F}\ \boldsymbol{s}^2\ \boldsymbol{s} \otimes \boldsymbol{u}\ n(t-\tau) - n(t)\sin\phi\ ...]. \tag{7}$$

The operator $\otimes$ denotes the element-wise products of vectors. The basic physical terms in $\boldsymbol{\theta}(\boldsymbol{s}, \boldsymbol{u}, \boldsymbol{F})$ can include constant, polynomial terms of $\boldsymbol{s}, \boldsymbol{u}$, and $\boldsymbol{F}$, etc. In addition, the time delayed term and trigonometric terms of $\phi$ are also included based on cutting force projection. By constructing the candidate set $\boldsymbol{\theta}(\boldsymbol{s}, \boldsymbol{u}, \boldsymbol{F})$, nonlinearities of machining dynamics can be introduced through the nonlinear combinations of state variables, process parameters, and cutting forces.

Then, the constructed candidate set $\boldsymbol{\theta}(\boldsymbol{s}, \boldsymbol{u}, \boldsymbol{F})$ is used to expand the function space for the unknow functions in Eq. (5). In this paper, two important assumptions are made. The first assumption is that all functions $f(\boldsymbol{s}, \boldsymbol{u}, \boldsymbol{F})$ in Eq. (5) live in the affine functional space expanded by $\boldsymbol{\theta}(\boldsymbol{s}, \boldsymbol{u}, \boldsymbol{F})$. That is, a cutting mechanics-based learning function space can be established as

$$f(\boldsymbol{s}, \boldsymbol{u}, \boldsymbol{F}) = \boldsymbol{\theta}(\boldsymbol{s}, \boldsymbol{u}, \boldsymbol{F}) \cdot \boldsymbol{\xi}, \tag{8}$$

where $\boldsymbol{\xi} = [\xi_1 \ \xi_2 \ \cdots \ \xi_P]^T \in \mathbb{R}^{P \times 1}$ represents the vector of the coefficients. The second assumption is that physical terms in the right-hand side of Eq. (8) are sparse in predefined function space, namely, the number of active terms (terms with nonzero coefficient in $\boldsymbol{\xi}$) in $f(\boldsymbol{s}, \boldsymbol{u}, \boldsymbol{F})$ is much smaller than that in the candidate set $\boldsymbol{\theta}(\boldsymbol{s}, \boldsymbol{u}, \boldsymbol{F})$.

*2.2.2 Discover DAEs by Discrete Optimization-based Learning Algorithm*

A discrete optimization-based learning algorithm is presented to estimate the coefficients $\boldsymbol{\xi}$ in Eq. (8) for data-driven discovery of unknown DAEs. A physical term in $\boldsymbol{\theta}(\boldsymbol{s}, \boldsymbol{u}, \boldsymbol{F})$ can be either present or not in $f(\boldsymbol{s}, \boldsymbol{u}, \boldsymbol{F})$. A new binary variable vector $\boldsymbol{\gamma}$, where $\boldsymbol{\gamma} = [\gamma_1 \ \gamma_2 \ \cdots \ \gamma_P]^T \in \mathbb{B}^{P \times 1}$, is introduced to indicate the presence of each physical term $\theta_p$ in $\boldsymbol{\theta}(\boldsymbol{s}, \boldsymbol{u}, \boldsymbol{F})$. Each binary variable $\gamma_p$ in the Boolean domain $\mathbb{B}$ is

$$\gamma_p = \begin{cases} 1, & \text{if } f(\boldsymbol{s}, \boldsymbol{u}, \boldsymbol{F}) \text{ includes } \theta_p \\ 0, & \text{otherwise} \end{cases}, \quad \forall p \in [P]. \tag{9}$$

With the indicator vector $\boldsymbol{\gamma}$, a sparse linear regression model can be established as in Eq. (10) to minimize the sum of squared errors between the measurements $\dot{\boldsymbol{x}}$ and model predictions of $f(\boldsymbol{s}, \boldsymbol{u}, \boldsymbol{F})$ with regularizations.

$$\min_{\boldsymbol{\gamma}, \boldsymbol{\xi}} \|\dot{\boldsymbol{x}} - \boldsymbol{\Theta}(\mathbf{S}, \mathbf{U}, \mathbf{F}) \cdot (\boldsymbol{\gamma} \circ \boldsymbol{\xi})\|_2^2 + \lambda_0 \|\boldsymbol{\xi}\|_0 + \lambda_2 \|\boldsymbol{\xi}\|_2^2. \tag{10}$$

$\boldsymbol{\Theta}(\mathbf{S}, \mathbf{U}, \mathbf{F}) \in \mathbb{R}^{N \times P}$ denotes the augmented matrix consisting of values of candidate terms that are calculated by evaluating the candidate set $\boldsymbol{\theta}(\boldsymbol{s}, \boldsymbol{u}, \boldsymbol{F})$ on datasets $\mathbf{S}$, $\mathbf{U}$, and $\mathbf{F}$. By introducing the element-wise product $\boldsymbol{\gamma} \circ \boldsymbol{\xi}$, Eq. (10) enables to control the presence of active physical terms in $\boldsymbol{\theta}(\boldsymbol{s}, \boldsymbol{u}, \boldsymbol{F})$ for the purpose of discovering parsimonious function $f(\boldsymbol{s}, \boldsymbol{u}, \boldsymbol{F})$. Two regularization terms are introduced. The number of active physical terms is implicitly controlled by the $l_0$-norm in the penalty $\lambda_0 \|\boldsymbol{\xi}\|_0$, which constraints the number of nonzero entries of $\boldsymbol{\xi}$. Another $l_2$-norm of $\boldsymbol{\xi}$, $\lambda_2 \|\boldsymbol{\xi}\|_2^2$, is introduced to reduce the effects of noisy process measurements. The penalty weights $\lambda_2$ and $\lambda_0$ are hyperparameters that control the strength of regularizations.

The sparse regression problem in Eq. (10) can be equivalently reformulated as a discrete optimization formulation as follows:

$$\min_{\gamma,\xi} \| \dot{x} - \Theta(S,U,F) \cdot \xi \|_2^2 + \lambda_2 \| \xi \|_2^2 \tag{11a}$$

$$s.t. \quad -M\gamma \le \xi \le M\gamma, \tag{11b}$$

$$\langle \gamma, e \rangle = k, \tag{11c}$$

$$\gamma \in \mathbb{B}^{P \times 1}, \xi \in \mathbb{R}^{P \times 1}. \tag{11d}$$

The $\ell_0$-norm and the relationship between $\xi$ and $\gamma$ in the objective function are converted to the constraints as shown in Eqs. (11b) - (11c), and the inequalities in Eq. (11b) indicate the element-wise relationship. When $\gamma_p = 0$, the $p$-th term in $\theta(s,u,F)$ is not in $f(s,u,F)$. As such, both sides of Eq. (11b) are equal to zero, leading to $\xi_p = 0$. When $\gamma_p = 1$, $\xi_p$ is bounded by a range $[-M, M]$, where $M$ is a constant number. $[-M, M]$ is the predefined lower and upper bound interval for coefficient $\xi$. The number of active physical terms is controlled by the inner product of $\gamma$ and $e$ in Eq. (11c), where $e \in \mathbb{R}^{P \times 1}$ is a vector with all entries set to 1.

To solve the discrete optimization problem in Eq. (11), a two-stage optimization algorithm presented in [22] is applied to determine $\gamma$ and $\xi$ separately. First, the column-wise normalization of data matrices $\Theta(S,U,F)$ and vector $\dot{x}$ is conducted to decrease the impacts of noise and physical scale of the data. The optimal term combination, $\gamma^*$, will be determined by solving Eq. (11) using column-wise normalized data of $\Theta(S,U,F)$ and $\dot{x}$. Then, the coefficients $\xi^*$ on the original data scale can be identified by a least square algorithm using the original data matrices $\Theta(S,U,F)$ and $\dot{x}$. The column corresponding to nonzero value in $\gamma^*$ is used for $\xi^*$ estimation.

Thus, the discovered governing DAE for displacement $\dot{x}$ can be given as follows:

$$\dot{x} = \theta(s,u,F) \cdot (\gamma^* \circ \xi^*). \tag{12}$$

The above algorithmic procedure is repeated to discover all the unknown DAEs in Eq. (5). In addition, chatter stability analysis can be conducted from both frequency and/or time domains by using the discovered governing DAEs of machining dynamics.

## 3 Simulation Experiments

Experimentally verified high-fidelity time domain simulation of milling is used to generate dynamic process data to validate the proposed CMML modeling method, in terms of its accuracy, efficiency, and robustness. All simulation experiments are performed on a mobile workstation with Intel(R) Xeon(R) W-10885M CPU @ 2.40GHz, 128 GB memory, and 64-bit Windows 11 Pro operating system for workstations.

### 3.1 Simulation Experiment Settings

*Time Domain Simulation of Milling*. Experimentally verified time domain simulator of milling in [1] is used for dynamic process data generation. This sim-to-real simulator enables the high accuracy numerical solution of the second order and coupled DDEs for milling dynamics by forward Euler method and has shown good agreement with physical machine tools [13-15]. It is well suited to incorporate the inherent complexities of milling dynamics, including complicated tool geometries (e.g., runout, different radii, number of the cutter teeth, non-uniform teeth spacing, and variable helix angles) and the nonlinearity that occurs due to interrupted cutting. Two case studies of governing equations discovery for milling dynamics are presented with increasing complexities:

- *Case I*: Milling dynamics with only time delayed effect in the cutting force model.
- *Case II*: Milling dynamics with time delayed effect, process damping, and edge force in the force model.

The above two milling cases refer to the most representative milling dynamics models that are dominantly used for decades of machining research. The use of these cases can demonstrate the general applicability of CMML method. It is noted that since turning dynamics can be regarded as a special case of milling dynamics by assuming constant immersion angle, the case study of turning is not presented in this paper. In both cases of milling, time series data is generated by time domain simulation. Gaussian noise is added to each measurable variable, namely the displacement, velocity, acceleration, and cutting forces as shown in Eq. (13):

$$\boldsymbol{S}^{noise} = \boldsymbol{S} + r\sigma_S \boldsymbol{\epsilon}. \tag{13}$$

$\boldsymbol{S} \in \mathbb{R}^{N \times 1}$ is an arbitrary column of data matrices $\boldsymbol{S}$, $\dot{\boldsymbol{S}}$, and $\boldsymbol{F}$. $\sigma_S$ is the standard deviation of the data column. $\boldsymbol{\epsilon} \in \mathbb{R}^{N \times 1}$ is the added noise vector with each entry drawn from a standard normal distribution $\mathcal{N}(0,1)$. $r$ is the noise ratio, where $r \in \{0, 0.01\%, 0.1\%, 1\%, 10\%, 50\%, 100\%\}$. Larger noise ratios indicate larger noises in data. A total of seven noise ratios are used to validate the proposed CMML modeling method.

*Parameter Settings of CMML*. The setting of three parameters of CMML are given as follows. The number of active terms $k$ is set to be the exact number of physical terms in the milling dynamics model, to test the capability of exact model discovery by CMML. The $\ell_2$-norm weight $\lambda_2$ is set to be 100 for dealing with data under large noise. The $M$ value for the bounds is set to be 1000. The hyperparameter tuning procedure for parameters $\lambda_2$ and $k$ by cross validation in [22] can also be applied. The discrete optimization problem in Eq. (11) is solved by branch-and-cut algorithm in modern discrete optimization solver CPLEX with version 20.1.

*Performance Metric.* To show the capability of CMML for exact governing equations discovery, the metric on the number of exactly discovered DAEs in Eq. (5) is defined as:

$$A := \sum_{i=1}^{2p+2q+2} 1_{\gamma_i^* = \gamma_i^\dagger} = \begin{cases} 1, & \text{if } \boldsymbol{\gamma}_i^* = \boldsymbol{\gamma}_i^\dagger \\ 0, & \text{if } \boldsymbol{\gamma}_i^* \neq \boldsymbol{\gamma}_i^\dagger \end{cases}. \tag{14}$$

$\boldsymbol{\gamma}_i^*$ indicates the active terms of the $i$-th DAE discovered by CMML, and $\boldsymbol{\gamma}_i^\dagger$ indicates the ground truth terms in the given governing equations of the studied milling system. Note here once the two algebraic equations for $F_t$ and $F_n$ are discovered, $F_x$ and $F_y$ will be determined. As such, in total $2p + 2q + 2$ DAEs need to be discovered by CMML. Besides this metric, the coefficient accuracy for each term is also reported.

### 3.2 Case I: Milling with Time Delayed Effect in Forces

*3.2.1 Exact Milling Governing Equations of Case I*

In this case study, the milling system with two orthogonal directions and a single mode on each direction is considered, namely $p = 1$ and $q = 1$. The force models only include regenerative time-delayed effect as shown in Eq. (15).

$$m_x \ddot{x}(t) + c_x \dot{x}(t) + k_x x(t) = F_x(t), \tag{15a}$$
$$m_y \ddot{y}(t) + c_y \dot{y}(t) + k_y y(t) = F_y(t). \tag{15b}$$

Cutting forces on the $x$ and $y$ directions are

$$F_x = -F_t \cos(\phi) + F_n \sin(\phi), \tag{15c}$$
$$F_y = F_t \sin(\phi) + F_n \cos(\phi). \tag{15d}$$

The tangential and normal force models are

$$F_t(t) = k_t b \big(f_t \sin \phi + n(t - \tau) - n(t)\big), \tag{15e}$$
$$F_n(t) = k_n b \big(f_t \sin \phi + n(t - \tau) - n(t)\big). \tag{15f}$$

$k_t$ and $k_n$ are cutting force coefficients for tangential and normal directions, $f_t$ is feed per tooth and $f_t \sin(\phi) + n(t - \tau) - n(t)$ is instantaneous chip thickness.

To validate the proposed CMML modeling method, the DDEs in Eq. (15) are further equivalently represented as a set of DAEs for the studied milling system in Eq. (16).

$$\dot{x} = v_x, \tag{16a}$$

$$\dot{v}_x = \frac{-c_x v_x - k_x x + F_x}{m_x}, \tag{16b}$$

$$\dot{y} = v_y, \tag{16c}$$

$$\dot{v}_y = \frac{-c_y v_y - k_y y + F_y}{m_y}, \tag{16d}$$

$$F_t = k_{tc} b \big( f_t \sin \phi + n(t - \tau) - n(t) \big), \tag{16e}$$

$$F_n = k_{nc} b \big( f_t \sin \phi + n(t - \tau) - n(t) \big), \tag{16f}$$

$$F_x = -F_t \cos(\phi) + F_n \sin(\phi), \tag{16g}$$

$$F_y = F_t \sin(\phi) + F_n \cos(\phi). \tag{16h}$$

Note that only Eqs. (16a) - (16f) are the six DAEs that need to be discovered by CMML. Eqs. (16g) - (16h) will be determined once Eqs. (16e) – (16f) are determined.

To discover the DAEs in Eqs. (16a) - (16f), the second-order polynomials are used to construct candidate terms in candidate set $\boldsymbol{\theta}(\boldsymbol{s}, \boldsymbol{u}, \boldsymbol{F})$ of CMML. By assuming zero cross talk between $x$ and $y$ directions, $\boldsymbol{s} = \{x, v_x\}$, $\boldsymbol{u} = \{b\}$, and $\boldsymbol{F} = \{F_x\}$ are used for discovering Eqs. (16a) - (16b); similarly, $\boldsymbol{s} = \{y, v_y\}$, $\boldsymbol{u} = \{b\}$, and $\boldsymbol{F} = \{F_y\}$ are used for discovering Eqs. (16c) - (16d). For discovering Eqs. (16e) - (16f), let $\Delta n = n(t - \tau) - n(t)$ denote the time delayed term. Thus, $\boldsymbol{s} = \{\Delta n\}$, $\boldsymbol{u} = \{b, \sin \phi\}$ and $\boldsymbol{F} = \emptyset$ are used to construct $\boldsymbol{\theta}(\boldsymbol{s}, \boldsymbol{u}, \boldsymbol{F})$. $\Delta n$ can be calculated with $x$ and $y$ based on the time delayed effect and projection relationship of cutting mechanics. Note that the spindle speed is not necessary to be included into $\boldsymbol{u}$, for it is used to calculate the time delay $\tau$ as $\tau = 1/(\Omega N_t)$ with a given $\Omega$ in rps and the number of teeth $N_t$.

### 3.2.2 Time Domain Simulation Setting for Case I

The setting of time domain simulation for Case I is given as follows. The modal parameters of the symmetric structural dynamics used are: $m_x = m_y = 0.198$ kg, $k_x = k_y = 5 \times 10^6$ N/m, and $c_x = c_y = 19.91$ N s/m. The cutting force parameters are set as $K_s = 750 \times 10^6$ N/m$^2$, cutting force angle $\beta = 68$ degree. Thus, $k_t = K_s \sin(\beta) = 695,387,669.36$ N/m$^2$, $k_r = K_s \cos(\beta) = 280,954,790.02$ N/m$^2$. The cutter has a diameter of 20 mm and four equally spaced straight teeth. Up milling with 25% radial immersion and feed rate $f_t = 0.1$ mm/tooth is used. The time domain simulation is executed for 40 revolutions and each revolution is discretized into 1000 steps. A set of process parameters $\langle \Omega, b \rangle$ is sampled to obtain dynamic process data from stable and unstable cuts, where $\Omega \in \{4000, 6000, 8000, 10000, 12000\}$ rpm and $b \in \{2, 4, 6, 8, 10, 12\}$ mm. Notably, the transient state data, which contains rich information of milling dynamics, is used for training of CMML. Thus, the first 2000 data points are used for each combination of $\langle \Omega, b \rangle$. As the cutting depth $b$ is treated as a variable, all the six datasets from six cutting depth $b$ under one spindle speed are stack as a single dataset. Thus, in total 12000 data points for each spindle speed are used for training of CMML to discover the set of DAEs in Eq. (16).

### 3.2.3 Numerical Results and Analyses of Case I

The numerical results of CMML for Case I of milling dynamics under different noise ratios are displayed in Tables 1 and 2. Table 1 shows the number of exactly discovered DAEs, based on the $A$ value, defined in Eq. (14). The number 6 indicates that all 4 differential equations to describe structural dynamics and 2 algebraic equations to represent $F_n$ and $F_t$ are exactly discovered. Under $\Omega = 6000$ rpm, CMML modeling method can exactly discover all the six DAEs from noisy data with noise ratio up to 50% (grey cells). Correspondingly, the discovered DAEs are presented in Table 2. Compared to the exact governing equations, the deviation of coefficients for each term is small (less than 0.03% on average). In addition, the computing time of CMML is 7.67 seconds on average for discovering the set of DAEs in Eq. (16). This shows the effectiveness, robustness, and efficiency of the proposed CMML modeling method to discover governing equations of milling dynamics in Case I from highly noisy data.

Table 1: Number of correct DAEs ($A$ value) discovered by CMML for the milling dynamics in Case I (with only time delayed effect in force models).

| Noise ratio $r$ (%) | Spindle Speed $\Omega$ (rpm) | | | | |
|---|---|---|---|---|---|
| | 4000 | 6000 | 8000 | 10000 | 12000 |
| Noise Free | 6 | 6 | 6 | 6 | 6 |
| 0.01% | 6 | 6 | 6 | 6 | 6 |
| 0.1% | 6 | 6 | 6 | 6 | 6 |
| 1% | 6 | 6 | 6 | 6 | 6 |
| 10% | 6 | 6 | 6 | 6 | 6 |
| 50% | 5 | 6 | 5 | 4 | 4 |
| 100% | 4 | 4 | 4 | 4 | 4 |
| 500% | 4 | 4 | 3 | 4 | 3 |
| 1000% | 2 | 2 | 2 | 2 | 2 |

Table 2: Discovered DAEs for the milling dynamics in Case I (with only time delayed effect in force model for $\Omega = 6000$ rpm in Table 1).

| Noise ratio $r$ (%) | Discovered DAEs |
|---|---|
| Noise Free (exact governing equations) | $\dot{x} = v_x$<br>$\dot{v}_x = -25266187.27x - 100.53v_x + 5.05F_x$<br>$\dot{y} = v_y$<br>$\dot{v}_y = -25266187.27y - 100.53v_y + 5.05F_y$<br>$F_t = -695387890.93\Delta nb + 69538.79 \sin \phi\, b$<br>$F_n = 280954945.06\Delta nb - 28095.50 \sin \phi\, b$ |
| 0.01% | $\dot{x} = v_x$<br>$\dot{v}_x = -25266192.30x - 100.53v_x + 5.05F_x$<br>$\dot{y} = v_y$<br>$\dot{v}_y = -25266191.61y - 100.53v_y + 5.05F_y$<br>$F_t = -695387669.36\Delta nb + 69538.81 \sin \phi\, b$<br>$F_n = 280954790.02\Delta nb - 28095.48 \sin \phi\, b$ |
| 0.1% | $\dot{x} = v_x$<br>$\dot{v}_x = -25266237.52x - 100.53v_x + 5.05F_x$<br>$\dot{y} = v_y$<br>$\dot{v}_y = -25266230.73y - 100.53v_y + 5.05F_y$<br>$F_t = -695385675.23\Delta nb + 69539.00\sin\phi\, b$<br>$F_n = 280953394.63\Delta nb - 28095.36\sin\phi\, b$ |
| 1% | $\dot{x} = v_x$<br>$\dot{v}_x = 25266679.36x - 100.55v_x + 5.05F_x$<br>$\dot{y} = v_y$<br>$\dot{v}_y = -25266620.77y - 100.50v_y + 5.05F_y$<br>$F_t = -695365734.00\Delta nb + 69540.87\sin\phi\, b$<br>$F_n = 280939440.76\Delta nb - 28094.17\sin\phi\, b$ |
| 10% | $\dot{x} = v_x$<br>$\dot{v}_x = -25270056.27x - 99.72v_x + 5.04F_x$<br>$\dot{y} = v_y$<br>$\dot{v}_y = -25270392.27y - 98.84v_y + 5.03F_y$<br>$F_t = -695166321.72\Delta nb + 69559.61\sin\phi\, b$<br>$F_n = 280799902.02\Delta nb - 28082.23\sin\phi\, b$ |
| 50% | $\dot{x} = v_x$<br>$\dot{v}_x = -25262721.75x - 75.51v_x + 4.70F_x$<br>$\dot{y} = v_y$<br>$\dot{v}_y = -25282589.36y - 66.14v_y + 4.43F_y$<br>$F_t = -694280044.90\Delta nb + 69642.87\sin\phi\, b$<br>$F_n = 280179729.87\Delta nb - 28029.19\sin\phi\, b$ |

Using the discovered DAEs with noise ratio from 0.00% (exact governing equations) to 50%, stability lobe diagrams are plotted in Figure 2 by using zero order method [24]. It is seen the stability lobes with noise ratio $r = 0.01\%, 0.1\%, 1\%$, and 10% almost overlap with that by the exact governing equations. This shows that the discovered DAEs are sufficiently accurate and robust for supporting stability analysis. As the noise ratio increases to 50%, the stability lobes (red virtual lines) deviate from the ground truth lobes (blue solid line).

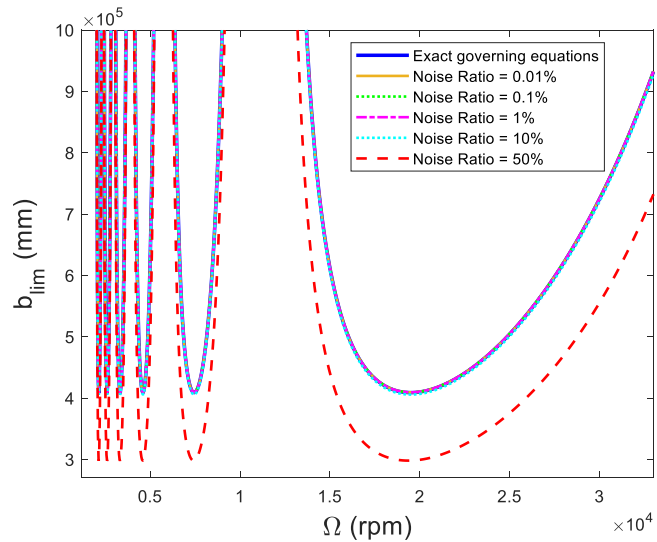

Figure 2: The stability lobe diagrams for $\Omega = 6000$ rpm in Table 2. The stability lobes under noises 0.01%, 0.1%, 1%, 10% almost perfectly overlap with stability lobes by exact governing equations.

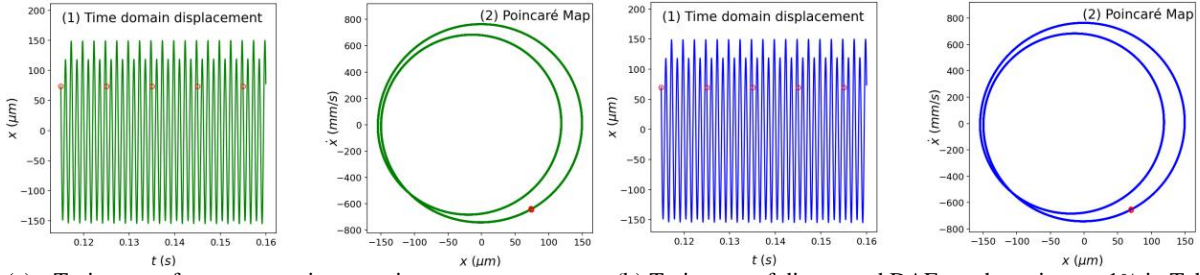

(a) Trajectory of exact governing equations  (b) Trajectory of discovered DAEs under noise $r = 1\%$ in Table 2.

Figure 3: Both (a) and (b) show the time domain displacement (on $x$ direction) and the Poincaré map of discovered equations in Table 2 using parameters $\Omega = 6000$ rpm and $b = 2$ mm. The single group of red points from once-per-tooth sampling show the cut is stable.

In addition, the time domain simulation results of the discovered DAEs are also presented. The time domain displacement and Poincaré maps obtained from the cut of $\Omega = 6000$ rpm and $b = 2$ mm are shown in Figure 3. The results for exact governing equations and equation discovered under noise ratio $r = 1\%$ are presented in Figures 3(a) and 3(b), respectively. The Poincaré map plots the velocity against the displacement. The single group of red points in Poincaré maps from once-per-tooth sampling indicate that the sampled points repeat, and the milling operation is stable. In general, both the time domain displacement and Poincaré map by the discovered DAEs align well with those by the exact governing equations.

### 3.3 Case II: Milling with Time Delayed Effect, Process Damping and Edge Force

This section presents the numerical results and analyses for milling dynamics in Case II with time delayed effect, process damping, and edge force. Nonlinear process damping is used to validate CMML for nonlinear DAEs discovery.

#### 3.3.1 Exact Milling Governing Equations and Simulation Data Generation of Case II

The exact governing equations used for Case II are the same as those in Case I for Eqs. (15a) - (15d), except for the following cutting force models:

$$F_t = k_{tc} b \big( f_t \sin\phi + n(t-\tau) - n(t) \big) + k_{te} b - C_t \frac{b}{V} \dot{n}^2, \tag{17a}$$

$$F_n = k_{nc} b \big( f_t \sin\phi + n(t-\tau) - n(t) \big) + k_{ne} b - C_n \frac{b}{V} \dot{n}^2. \tag{17b}$$

Instead of using the linear process damping force in [9], a nonlinear process damping term $-C\frac{b}{V}\dot{n}^2$ is informally introduced in the cutting force model in Eq. (17) for demonstrating the capability of CMML for nonlinear model discovery. For the introduced nonlinear process damping term $-C\frac{b}{V}\dot{n}^2$, $\dot{n}, b, V$, and $C$ are the normal velocity, chip width, cutting speed, and constant coefficient. For the cutting force coefficients $k_{tc}, k_{nc}, k_{te}$, and $k_{ne}$, subscripts $t$ and $n$ refer to the tangential and normal directions, $c$ and $e$ refer to the cutting and edge effects.

The simulation data is obtained from time domain simulation using the same parameter settings in Case I except for the follows: process damping coefficients $C_t = C_n = 1.4 \times 10^3$ N s/m², and edge force coefficients $k_{te} = k_{ne} = 2.5 \times 10^4$ N/m. Because of the introduced nonlinear process damping term, third-order polynomials of $s = \{\Delta n, \dot{n}\}$ and $u = \{b, \sin\phi\}$ are used by CMML for discovering the algebraic equations of cutting forces. $\Delta n$ and $\dot{n}$ can be calculated with $x, v_x, y$, and $v_y$ based on the time delayed effect and projection relationship of cutting mechanics. Other settings of CMML modeling method remain unchanged as in Case I.

*3.3.2 Numerical Results and Analyses of Case II*

The numerical results of CMML to discover the DAEs of milling dynamics in Case II under different noise ratios are summarized in Table 3, Table 4, and Figure 4. The number of exactly discovered DAEs for milling dynamics in Case II is presented in Table 3. It is seen that CMML can exactly discover all the six DAEs from noisy data with noise ratio up to 50%. The computing time of CMML is 9.23 seconds on average for discovering a set of DAEs of milling dynamics in Case II. In addition, it is seen a lower value of $\Omega$ can lead to better performance of CMML for exact DAEs discovery. This can be used for guiding experiment design as lower value of spindle speed is good for tool life and can reduce potential damage to machine tool.

In addition, Table 4 shows the discovered DAEs of $\Omega = 6000$ rpm under different noise ratios (see grey cells of Table 3). All physical terms including the nonlinear process damping are discovered correctly. In Figure 4, the time domain displacement and Poincaré maps are plotted for the exact governing equations and the DAEs obtained under noise ratio 1% in Table 3 with parameters $\Omega = 6000$ rpm and $b = 2$ mm. Similar conclusions can be made as in Case I. The stability lobe diagram is not presented due to no available methods to obtain the stability solutions of nonlinear governing equations. In general, the numerical results of Case II show that CMML can discover accurate nonlinear governing equations of machining dynamics from highly noisy data.

Table 3: Number of correct DAEs ($A$ value) discovered by CMML for the milling dynamics in Case II with time delayed effect, nonlinear process damping, and edge effect in the force models.

|                   | Spindle Speed $\Omega$ (rpm) | | | | |
|-------------------|------|------|------|-------|-------|
| Noise ratio $r$ (%) | 4000 | 6000 | 8000 | 10000 | 12000 |
| Noise Free        | 6    | 6    | 6    | 6     | 6     |
| 0.01%             | 6    | 6    | 6    | 6     | 6     |
| 0.1%              | 6    | 6    | 6    | 6     | 6     |
| 1%                | 6    | 6    | 6    | 6     | 6     |
| 10%               | 6    | 6    | 4    | 5     | 6     |
| 50%               | 6    | 6    | 2    | 2     | 4     |

| 100% | 4 | 4 | 3 | 3 | 3 |
| --- | --- | --- | --- | --- | --- |
| 500% | 2 | 3 | 2 | 2 | 2 |
| 1000% | 2 | 2 | 2 | 2 | 2 |

Table 4: Discovered DAEs for the milling dynamics in Case II (with time delayed effect, nonlinear process damping, and edge effect in the force models for $\Omega = 6000$ rpm in Table 3).

| Noise ratio $r$ (%) | Discovered DAEs |
| --- | --- |
| Noise Free (exact governing equations) | $\dot{x} = v_x$<br>$\dot{v}_x = -25266187.27x - 100.53v_x + 5.05F_x$<br>$\dot{y} = v_y$<br>$\dot{v}_y = -25266187.27y - 100.53v_y + 5.05F_y$<br>$F_t = 25000.00b - 695387890.92\Delta nb$<br>$\quad +69538.79\sin\phi\, b - 222.82\dot{n}^2 b$<br>$F_n = 25000.00b + 280954945.06\Delta nb$<br>$\quad -28095.49\sin\phi\, b + 222.82\dot{n}^2 b$ |
| 0.01% | $\dot{x} = v_x$<br>$\dot{v}_x = -25266185.24x - 100.53v_x + 5.05F_x$<br>$\dot{y} = v_y$<br>$\dot{v}_y = -25266192.35y - 100.53v_y + 5.05F_y$<br>$F_t = 25000.44b - 695388128.39\Delta nb$<br>$\quad +69538.33\sin\phi\, b - 222.81\dot{n}^2 b$<br>$F_n = 24999.86b + 280954746.63\Delta nb$<br>$\quad -28095.88\sin\phi\, b + 222.82\dot{n}^2 b$ |
| 0.1% | $\dot{x} = v_x$<br>$\dot{v}_x = -25266166.88x - 100.54v_x + 5.05F_x$<br>$\dot{y} = v_y$<br>$\dot{v}_y = -25266237.93y - 100.53v_y + 5.05F_y$<br>$F_t = 25004.41b - 695390265.56\Delta nb$<br>$\quad +69534.22\sin\phi\, b - 222.80\dot{n}^2 b$<br>$F_n = 24998.61b + 280952960.82\Delta nb$<br>$\quad -28099.37\sin\phi\, b + 222.83\dot{n}^2 b$ |
| 1% | $\dot{x} = v_x$<br>$\dot{v}_x = -25265972.09x - 100.59v_x + 5.05F_x$<br>$\dot{y} = v_y$<br>$\dot{v}_y = -25266678.29y - 100.46v_y + 5.05F_y$<br>$F_t = 25044.14b - 695411637.29\Delta nb$<br>$\quad +69493.07\sin\phi\, b - 222.61\dot{n}^2 b$<br>$F_n = 24986.10b + 280935102.62\Delta nb$<br>$\quad -28134.27\sin\phi\, b + 222.96\dot{n}^2 b$ |
| 10% | $\dot{x} = v_x$<br>$\dot{v}_x = -25262914.86x - 99.77v_x + 5.04F_x$<br>$\dot{y} = v_y$<br>$\dot{v}_y = -25269532.49y - 98.42v_y + 5.03F_y$<br>$F_t = 25441.38b - 695625354.64\Delta nb$<br>$\quad +69081.56\sin\phi\, b - 220.80\dot{n}^2 b$<br>$F_n = 24861.03b + 280756520.68\Delta nb$<br>$\quad -28483.26\sin\phi\, b + 224.26\dot{n}^2 b$ |
| 50% | $\dot{x} = v_x$<br>$\dot{v}_x = -25227362.02x - 69.90v_x + 4.73F_x$<br>$\dot{y} = v_y$<br>$\dot{v}_y = -25249866.63y - 62.19v_y + 4.69F_y$<br>$F_t = 27206.92b - 696575209.50\Delta nb$<br>$\quad +67252.64\sin\phi\, b - 212.71\dot{n}^2 b$<br>$F_n = 24305.18b + 279962823.18\Delta nb$<br>$\quad -30034.33\sin\phi\, b + 230.01\dot{n}^2 b$ |

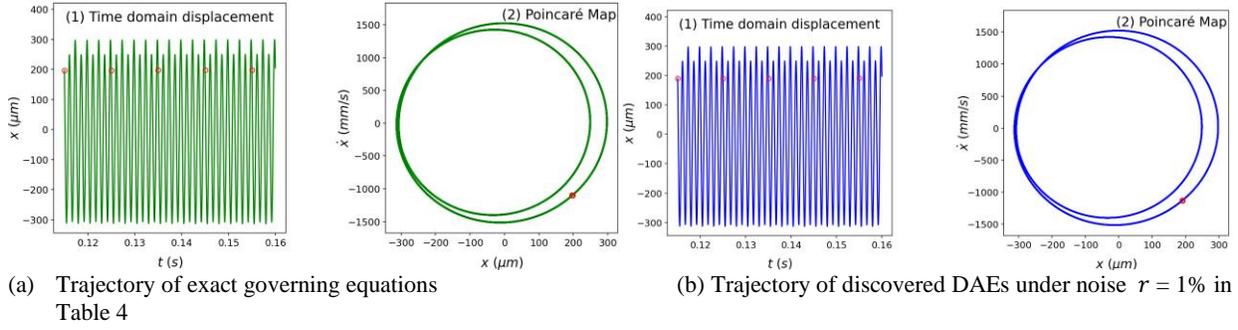

(a) Trajectory of exact governing equations

(b) Trajectory of discovered DAEs under noise $r = 1\%$ in Table 4

Figure 4: Both (a) and (b) show the time domain displacement (for $x$ direction) and the Poincaré map of discovered equations in Table 4 using parameters $\Omega = 6000$ rpm and $b = 2$ mm. The single group of red points from once-per-tooth sampling shows the cut is stable.

In summary, numerical results in Sections 3.2 and 3.3 demonstrate the effectiveness, accuracy, and robustness of the proposed CMML modeling method for discovering governing equations of machining dynamics. Under high noise levels, CMML can successfully discover the exact DAEs for machining dynamics in both Cases I and II with time delayed effect, process damping, and edge effect. This indicates that CMML has the potential to be used for advancing machining modeling in practice with the development of effective metrology systems.

## 4 Conclusions

This paper proposes a cutting mechanics-based machine learning (CMML) modeling method, that provides a new pathway to integrate existing physics in cutting mechanics and unknown physics in data, to discover governing equations of machining dynamics. In CMML, existing well-established physical principles of cutting mechanics are utilized at first to establish a general modeling structure to guide data-driven algorithm design. This is done through representing the second order, coupled, time delayed differential equations (DDEs) by a set of unknown differential algebraic equations. This representation strictly follows the known physics in both structural dynamics and cutting forces. Then, CMML can achieve data-driven discovery of these unknown differential algebraic equations by developing customized machine learning algorithm, including nonlinear learning function space design and discrete optimization-based training algorithm. Experimentally verified time domain simulation of milling is used to validate the proposed modeling method. Numerical results show CMML can discover the exact milling dynamics governing equations with time delayed effect, process damping, and edge force from noisy data.

Future work will focus on developing effective metrology systems to simultaneously measure the cutting force and at least one among the displacement, velocity and acceleration. The proposed CMML modeling method can therefore be used for discovery of nonlinear governing equations of machining dynamics from experimental data to advance machining modeling in practice, especially for nonlinear process damping models. In addition, developing efficient stability solution algorithms to accurately and fast solve nonlinear DDEs will also be considered.

**Acknowledgements**

The authors acknowledge support from the NSF Engineering Research Center for Hybrid Autonomous Manufacturing Moving from Evolution to Revolution (ERC-HAMMER) under Award Number EEC-2133630. This work was partially supported by the DOE Office of Energy Efficiency and Renewable Energy (EERE), under contract DE-AC05 00OR22725. The authors also gratefully acknowledge the AI Tennessee Initiative and Southeastern Advanced Machine Tools Network (SEAMTN) to partially support this research.